\documentclass[letterpaper, 10 pt, conference]{ieeeconf}  

\IEEEoverridecommandlockouts
\usepackage{times}
\usepackage{cite}
\usepackage{multicol}
\usepackage[bookmarks=true]{hyperref}

\usepackage{wrapfig}
\usepackage{booktabs}
\usepackage{caption}
\usepackage{multirow}
\usepackage{colortbl}
\usepackage{makecell}
\usepackage{amsmath} 
\usepackage{amssymb}  
\usepackage{pifont}
\newcommand{\xmark}{\ding{55}}%
\usepackage[]{changes}
\presetkeys%
    {todonotes}%
    {inline,backgroundcolor=yellow}{}
\usepackage{chngcntr}
\usepackage{siunitx}
\usepackage{tikz}
\usepackage{adjustbox}
\usepackage{placeins}
\usepackage{url}
\usepackage{subfigure}

\title{ViSA-Flow: Accelerating Robot Skill Learning via \\ Large-Scale Video Semantic Action Flow
}
\author{Changhe Chen$^{*1}$, Quantao Yang$^{*2}$, Xiaohao Xu$^{1}$, Nima Fazeli$^{1}$, and Olov Andersson$^2$ 
\thanks{This work was partially supported by the Wallenberg AI, Autonomous Systems and Software Program (WASP) funded by the Knut and Alice Wallenberg Foundation.}
\thanks{$*$These authors contributed equally.}
\thanks{$^{1}$University of Michigan. {\tt (changhec@umich.edu)}.}
\thanks{$^{2}$Division of Robotics, Perception and Learning (RPL), KTH Royal Institute of Technology, Sweden. {\tt (quantao@kth.se)}.
}%
}%

\begin{document}
\maketitle
\vspace{-0.0cm}



\begin{abstract}
One of the central challenges preventing robots from acquiring complex manipulation skills is the prohibitive cost of collecting large-scale robot demonstrations. In contrast, humans are able to learn efficiently by watching others interact with their environment. To bridge this gap, we introduce \textit{semantic action flow} as a core intermediate representation capturing the essential spatio-temporal manipulator-object interactions, invariant to superficial visual differences. We present ViSA-Flow, a framework that learns this representation self-supervised from unlabeled large-scale video data. First, a generative model is pre-trained on semantic action flows automatically extracted from large-scale human-object interaction video data, learning a robust prior over manipulation structure. Second, this prior is efficiently adapted to a target robot by fine-tuning on a small set of robot demonstrations processed through the same semantic abstraction pipeline. We demonstrate through extensive experiments on the CALVIN benchmark and real-world tasks that ViSA-Flow achieves state-of-the-art performance, particularly in low-data regimes, outperforming prior methods by effectively transferring knowledge from human video observation to robotic execution. Videos are available at \url{https://visaflow-web.github.io/ViSAFLOW}.
 
\end{abstract}

\IEEEpeerreviewmaketitle


\section{Introduction}
Robot imitation learning has achieved remarkable success in enabling robots to acquire complex manipulation skills, ranging from basic object manipulation\cite{hussein2017imitation, gao2024prime} to intricate assembly procedures\cite{chi2023diffusion}. However, the scalability of imitation learning approaches is fundamentally limited by the need for extensive, carefully curated robot datasets that are costly to collect. This has become a critical bottleneck in developing robots capable of performing diverse real-world tasks.

In contrast, humans demonstrate an extraordinary ability to learn new skills by observing others. From in-person observation to videos, humans naturally focus on semantically relevant components. For instance, when learning tennis, we naturally attend to the player's body movements, racquet handling techniques, and ball trajectories, while effectively filtering out irrelevant background information. This selective attention to meaningful elements enables efficient skill acquisition and transfer. The vast repository of publicly available videos on the internet similarly represents an untapped resource for robot learning, offering diverse demonstrations of human skills across countless domains. However, effectively leveraging this resource requires addressing several key challenges, particularly in bridging the gap between human demonstrations in unconstrained videos and robot execution in the real world.



\begin{figure}[t!]
    \centering
    \includegraphics[width=\linewidth]{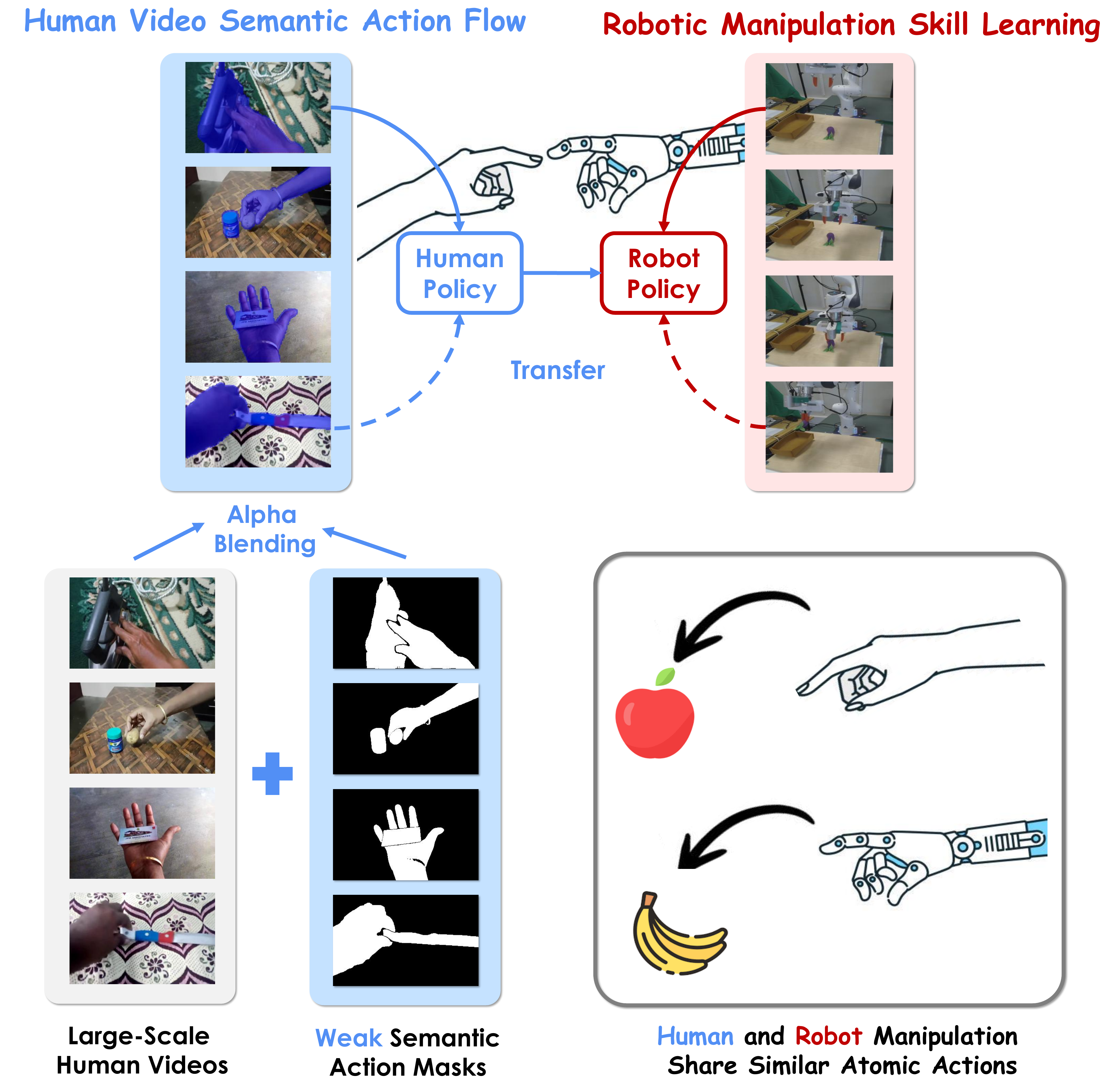}
      \caption{
      Humans and robots often share underlying atomic actions for similar tasks. 
      Our framework leverages large-scale, unlabeled human videos by extracting weakly supervised semantic action flow priors (ViSA-Flow). This knowledge is distilled into a human policy and efficiently transferred to learn a corresponding robot policy.
      }
      \label{fig: teaser}
      \vspace{-\baselineskip}
\end{figure}

Recent research\cite{bahl2022human, zeng2024learning, li2024grmg} has investigated how robots can acquire skills by directly observing unstructured human videos. These methods have shown promising generalization, enabling robots to adapt to new tasks beyond their training data. However, most existing approaches\cite{wen2024anypointtrajectorymodelingpolicy, xu2024flowcrossdomainmanipulationinterface, yuan2024generalflowfoundationaffordance} rely heavily on motion flow as a conditional input for policy learning. While effective in some settings, this low-level representation often overlooks the higher-level semantic cues that humans naturally attend to when learning new skills.
In real-world scenarios, when humans acquire a skill, we rarely process the entire visual scene indiscriminately. Instead, we focus selectively on the interaction between the hand (or arm) and the relevant object, while disregarding irrelevant background elements or distractions. Mimicking this selective attention mechanism could make robot learning from videos more efficient and robust.

Drawing inspiration from this observation, we propose a novel framework that learns robot skills by extracting temporally consistent semantic action flows from large-scale human manipulation videos (Fig.~\ref{fig: teaser}). Unlike prior works that condition policy learning on motion flow, our approach leverages semantic representations—capturing object interactions, body poses, and motion patterns—that can be consistently transferred from human demonstrations to robotic actions. By focusing on these meaningful semantic structures, our method enables more efficient and generalizable skill learning from videos. Our key contributions are threefold: 
\begin{enumerate}
    \item We propose \textbf{ViSA-Flow}, a framework for pre-training generative policies using large-scale \textbf{Vi}deo \textbf{S}emantic \textbf{A}ction \textbf{Flow},  capturing spatio-temporal manipulator-object interactions from diverse human video demonstrations. This enables efficient knowledge transfer from large-scale human video data to robotic manipulation policies and offers a new perspective on representing action sequences for cross-domain transfer.

    \item We refine the pretrained policy using robot-specific semantic actions from few expert demonstrations by tracking hand-object interactions in both human videos and robot data, enabling robust semantic alignment for improved policy adaptation.
    
    \item We evaluate ViSA-Flow in both simulated and real-world robotic manipulation tasks, demonstrating substantial performance improvements over SOTA baselines, particularly in low-data regimes. 
\end{enumerate}


\section{Related Work}
\label{sec:citations}
\label{sec:related_work}
\textbf{Visual Imitation Learning.}
Recent advancements\cite{tian2024predictiveinversedynamicsmodels, bu2024closedloopvisuomotorcontrolgenerative, yang2025s, mees2022matterslanguageconditionedrobotic, black2023susie} in visual feature-based imitation learning have significantly improved the efficiency, generalization of learning from visual demonstrations. VIEW\cite{jonnavittula2025viewvisualimitationlearning} introduces a trajectory segmentation approach that extracts condensed prior trajectories from demonstrations, allowing robots to learn manipulation tasks more efficiently. Similarly, K-VIL\cite{Gao_2023} enhances efficiency by extracting sparse, object-centric keypoints from visual demonstrations, reducing redundancy and improving learning speed. Beyond efficiency, generalization remains a critical challenge, particularly in adapting to diverse visual environments. Stem-OB\cite{hu2024stemobgeneralizablevisualimitation} addresses this issue by leveraging diffusion model inversion to suppress low-level visual differences, improving robustness against variations in lighting and texture. 
In addition, goal-oriented approaches have been developed to improve policy learning and adaptation. Visual hindsight self-imitation learning\cite{Kim_2024} introduces hindsight goal re-labeling and prototypical goal embedding, enhancing sample efficiency in vision-based tasks. 


\textbf{Video-Based Robot Learning.}
Recent advancements\cite{brohan2023rt2, reuss2024multimodaldiffusiontransformerlearning, zhou2024languageconditionedimitationlearningbase, li2024visionlanguagefoundationmodelseffective} in robot learning have demonstrated the effectiveness of large-scale video datasets for pre-training models and improving generalization. Methods such as Time-Contrastive Networks (TCN)\cite{sermanet2018timecontrastivenetworksselfsupervisedlearning} have pioneered the extraction of temporally consistent features to align human demonstrations with robot actions. Building on this foundation, video pretraining\cite{baker2022videopretrainingvptlearning} has shown that large-scale video data can be used to pretrain robust visual representations for downstream manipulation tasks.
More recent works\cite{wu2023unleashinglargescalevideogenerative} have further leveraged large-scale video datasets to enhance manipulation performance. 
Similarly, Vid2Robot \cite{jain2024vid2robotendtoendvideoconditionedpolicy} directly translates video demonstrations into robot actions using cross-attention for alignment.
\cite{li2024grmg} highlights the potential of leveraging partially-annotated data to enhance robot policy learning by integrating multi-modal information. \cite{lepert2503phantom} transforms human video demonstrations into robot-compatible observation-action pairs by inpainting the human arm, and overlaying a rendered robot to achieve visual domain alignment.
Beyond task-specific learning, Ye et al. \cite{ye2025video2policyscalingmanipulationtasks} explored scaling up robot learning via Internet videos, investigating how web-scale human video datasets can enhance policy learning efficiency. 

\textbf{Flow-Guided Imitation Learning.}
A growing body of work has explored representing manipulation trajectories in more flexible and generalizable forms to enhance policy learning. Wen et al.\cite{wen2024anypointtrajectorymodelingpolicy} introduced Any-point Trajectory Modeling (ATM), which allows policies to query and generate trajectory states at arbitrary temporal points. This continuous-time representation enables efficient interpolation and improved temporal flexibility, reducing dependence on fixed-horizon action sequences. 
Bharadhwaj et al.\cite{bharadhwaj2024track2actpredictingpointtracks} proposed Track2Act, which leverages point tracks extracted from Internet videos to learn generalizable manipulation skills. By grounding policies in persistent motion tracks, their method facilitates robust transfer across embodiments and visual domains without requiring paired robot demonstrations.

Another emerging line of research focuses on using \emph{flow}-based representations as a domain-agnostic interface for manipulation. Xu et al.\cite{xu2024flowcrossdomainmanipulationinterface} introduced a method that predicts dense optical flow fields as an intermediate representation, enabling the transfer of skills across domains by decoupling perception from control. Extending this idea, Yuan et al.\cite{yuan2024generalflowfoundationaffordance} proposed general flow as a unified affordance representation for scalable robot learning, demonstrating its capacity to generalize across object categories and manipulation tasks. Flow-based interfaces effectively capture motion intent while enabling cross-domain consistency. Ren et al.\cite{ren2025motiontracksunifiedrepresentation} introduced Motion Tracks, an agent-agnostic trajectory representation for few-shot human-to-robot transfer, enabling rapid adaptation to new tasks with minimal robot-specific data. 

All these works leverage dense optical flow or motion tracks as intermediate representations, enabling skill transfer across domains. These approaches have shown strong results using non-robot videos to improve robot policies. Our method differs in the type of representation: rather than relying on flow fields, we use weak hand–object segmentation masks amplified by temporal tracking (Sec.~\ref{sec:method}), aiming to capture higher-level semantic interactions.

\section{Method}
\label{sec:method}

\begin{figure*}[t!]
    \centering
    \includegraphics[width=1.0\linewidth]{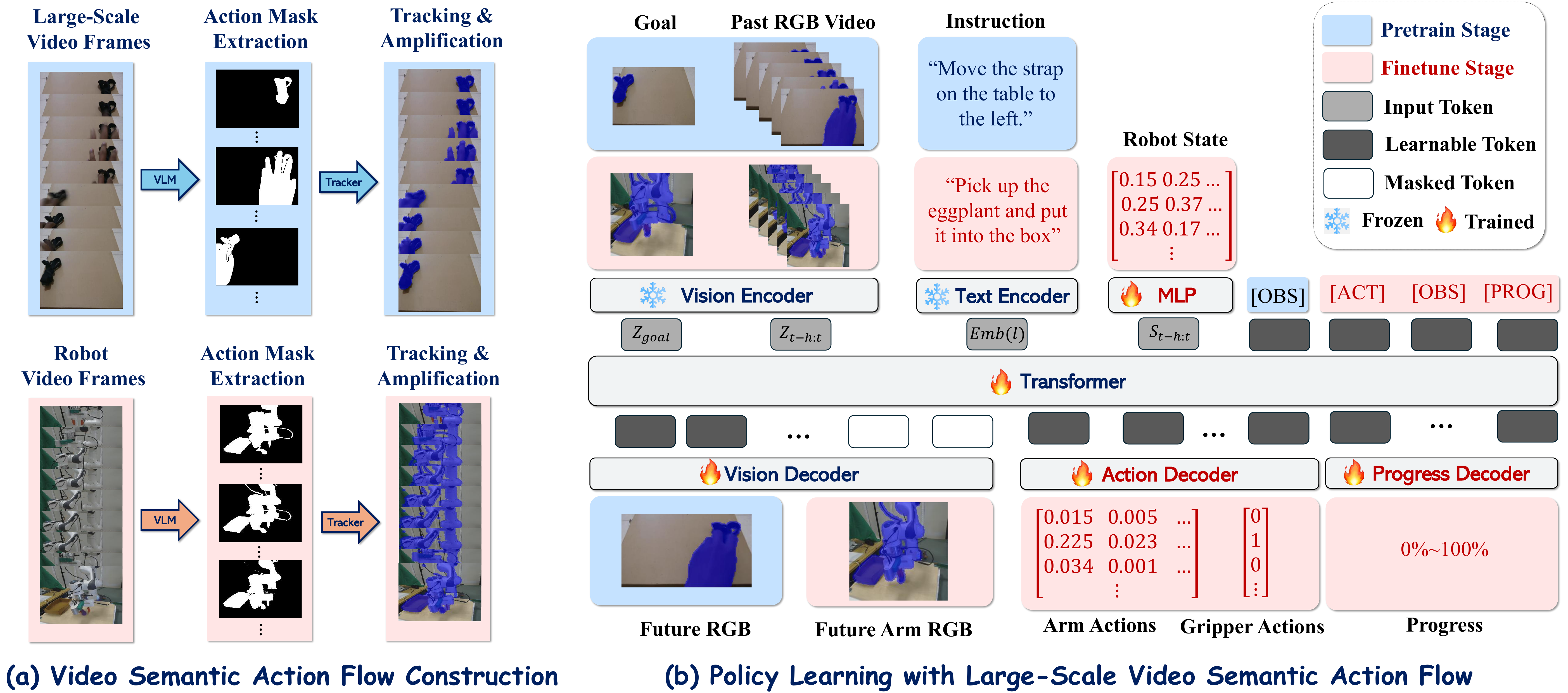}      \caption{\textbf{ViSA-Flow Architecture and Policy Learning Framework.} (\textbf{a}) During pretraining, hand-object interaction masks are extracted from large-scale video frames and amplified via tracking to generate semantic flow representations. (\textbf{b}) In the finetuning stage, a multi-modal Transformer architecture conditions on the goal image, a sequence of RGB observation frames enhanced with pre-trained ViSA-Flow, language instructions and robot state. The Transformer predicts future visual states, low-level robot actions, and task progress using dedicated decoders.}
      \vspace{-0.mm}
      \label{fig: visa_pipeline}
      \vspace{-\baselineskip}
\end{figure*}


Our approach facilitates learning robot manipulation policies from limited \textit{target-domain} data by leveraging knowledge distilled from large-scale \textit{source-domain} (human) videos. This is achieved through the introduction and utilization of \textbf{Video Semantic Action Flow (ViSA-Flow)}, a structured intermediate representation designed for cross-domain transfer. We first formulate the conceptual properties of ViSA-Flow and motivate its suitability for transfer learning, then detail its concrete implementation via our two-stage learning framework.

\subsection{Problem Definition}
\label{sec:problem_definition}
Our objective is to pretrain a policy model $\pi_\theta$ by utilizing human-object interactions from a large dataset of human manipulation videos, $D_v = \{v_i\}^M$. This pretraining aims to facilitate learning on a target robotic task using only a small dataset of robot demonstrations, $D_{\tau} = \{\tau_j\}^N$, where $N \ll M$.
The target task involves controlling a robot based on language instructions, observations, and proprioceptive state. We define the robot's observation space as $O$, its proprioceptive state space as $S$, and its action space as $A$. Given a language instruction $l$, our goal is to learn a policy $\pi_\theta(a_t | l, o_{t-h:t}, s_{t-h:t})$ that outputs an action $a_t \in A$ based on the instruction $l$, a history of recent observations $o_{t-h:t} \in O$, and recent states $s_{t-h:t} \in S$. This policy is learned primarily by imitating the demonstrations in $D_{\tau}$, leveraging the pretraining from $D_v$.


\subsection{ViSA-Flow Representation}
\label{sec:visaflow_representation}
We propose ViSA-Flow as an intermediate representation $z_t \in Z_{\text{ViSA-Flow}}$ obtained by mapping an observation $o_t$ and context $l$ through a function $f: O \times L \rightarrow Z_{\text{ViSA-Flow}}$. We design $Z_{\text{ViSA-Flow}}$ to preserve task-relevant interactions while mitigating domain-specific nuisance factors, enabling cross-domain transfer. 

\paragraph{Semantic Entity Grounding.} Given the initial observation frame $o_0$ and context $l$, we utilize a pre-trained Vision-Language Model (VLM) to ground textual descriptions of the manipulator (e.g., `hand', `gripper') and task-relevant objects (e.g., `red block') identified from $l$. A segmentation model (e.g., SAM\cite{kirillov2023segment}) then generates initial segmentation masks for these grounded entities, including manipulators and objects, \textit{i.e.,} $\{m_{M,0}, m_{O_k,0}\}$.

\paragraph{Hand-Object Interaction Tracking.} Due to the instability of semantic segmentation across sequential frames,  we propose tracking the correctly segmented hand-object interaction mask over time. Specifically, we instantiate a robust point tracker (e.g., CoTracker\cite{karaev2024cotracker3simplerbetterpoint}) with points densely sampled within the initial masks. The tracker estimates the 2D image trajectories $P_t = \{p_{j,t}\}_{j=0}^J$ for these points across the sequence $\{o_t\}_{t=0}^T$. These trajectories $P_t$ represent the extracted raw flow information.

\paragraph{Flow-Conditioned Feature Encoding.} To produce the final VISA-Flow  representation $z_t$, we encode the flow information $P_t$  into a rich feature vector while  retaining  visual context. We first apply a perceptual enhancement process directly on the raw observation frame $o_t$. Using tracked point trajectories $P_t$, we generate a spatially-localized amplification mask $M_t(x,y)$ with parameterized radius $r$ around each tracker coordinate:
\begin{equation}
\label{eq:visaflow_mask}
\begin{aligned}
M_t(x,y) &= \max_{p \in P_t} \mathbf{1}\!\bigl(\lVert(x,y) - p\rVert_2 \le r\bigr).
\end{aligned}
\end{equation}
This mask modulates pixel intensities by an amplification factor $\alpha$ within these regions of interest, while maintaining contextual information elsewhere. The resulting perceptually-enhanced frame exhibits selective luminance amplification at interaction-critical regions. This pre-processed frame is then passed through a pre-trained vision encoder $\phi$ (e.g., MAE\cite{he2021maskedautoencodersscalablevision}) which is frozen during policy learning, transforming the flow-highlighted observations into our implemented ViSA-Flow representation $z_t$:
\begin{equation}
\label{eq:visaflow_representation}
\begin{aligned}
z_t      &= \phi\!\bigl(o_t \odot [\,1 + \alpha\,M_t\,]\bigr).
\end{aligned}
\end{equation}

This implementation aims to focus on tracked semantic entities and modulating features accordingly.

\subsection{Policy Learning through ViSA-Flow Representation}
\label{sec:learning_framework_architecture}
Our learning framework leverages the extracted ViSA-Flow representations $z_t$ within a two-stage pre-training and fine-tuning scheme, implemented using a transformer architecture, denoted $g_\psi$ (parameters $\psi$), inspired by prior work such as GR-1\cite{wu2023unleashinglargescalevideogenerative}.

\paragraph{Model Architecture.} A transformer $g_\psi$ is designed to process multimodal sequences for both generative prediction and policy inference shown in Fig.~\ref{fig: visa_pipeline}. Its input is a sequence formed by concatenating tokens representing various modalities and special learnable query tokens. Primary input modalities include language instruction embeddings $\text{Emb}(l)$ (e.g., from CLIP\cite{radford2021learningtransferablevisualmodels}), the sequence of recent ViSA-Flow representations $\{z_{t-h}, ..., z_t\}$ encoding flow-conditioned visual features (Sec.~\ref{sec:visaflow_representation}), the sequence of proprioceptive states $\{s_{t-h}, ..., s_t\}$ (processed via linear embeddings), and potentially tokens representing a goal state $z_{goal}$. Added to these are special query tokens: an \texttt{[ACT]} token for action prediction and multiple \texttt{[OBS]} tokens for predicting future ViSA-Flow states. Standard positional embeddings are added to this combined sequence to encode temporal order before processing by the transformer blocks. 
The output embeddings corresponding to the query tokens are then directed to task-specific heads; notably, the \texttt{[ACT]} token's output yields the action chunk prediction $\hat{a}_{t+1:t+k}$, while the \texttt{[OBS]} tokens' outputs yield predictions $\hat{z}_{t+1:t+n}$ for future states.

\paragraph{Stage 1: Pre-training -- Learning ViSA-Flow Dynamics Prior.} 
Using the large-scale human video dataset $D_v$, we pre-train $g_\psi$ to model the dynamics within the ViSA-Flow space. For each sequence $v_i \in D_v$, we extract $\{z_{i,t}\}$ (Sec.~\ref{sec:visaflow_representation}). The model is trained to predict future representations $z_{t+1:t+n}$ based on past context $z_{\le t}$ and $l$, using the \texttt{[OBS]} query tokens. The objective is to minimize the prediction error, typically via Mean Squared Error (MSE):
\begin{equation}
    \mathcal{L}_{\text{pretrain}}(\psi) = \mathbb{E}_{v \sim D_v} \left[ || g_\psi(z_{\le t}, l)_{\text{[OBS]}} - z_{t+1:t+n} ||^2 \right].
    \label{eq:pretrain_loss_final}
\end{equation}
This stage yields pre-trained parameters $\psi_{\text{pre}}$, encoding a prior over interaction dynamics.

\paragraph{Stage 2: Fine-tuning -- Policy Adaptation.} 
Using the small-scale robot demonstration dataset $D_{\tau}$, we fine-tune the model, initialized with $\psi_{\text{pre}}$, to learn the target policy $\pi_\theta$ (where $\theta \subseteq \psi$). For each robot trajectory $\tau_j \in D_{\tau}$, we extract ViSA-Flow representations $\{z_{j,t}\}$ using the identical pipeline. The model is trained end-to-end with a multi-task objective combining action prediction and continued dynamics modeling:
\begin{equation}
\begin{split}
\mathcal{L}_{\text{finetune}}(\psi)=
\mathbb{E}_{\tau \sim D_{\tau}}\Bigl[
  \mathcal{L}_{\text{act}}\bigl(a_{t+1:t+k},\,\hat{a}_{t+1:t+k}\bigr)\\
  {}+\lambda_{\text{fwd}}\,
  \mathcal{L}_{\text{obs}}\bigl(z_{t+1:t+n},\,\hat{z}_{t+1:t+n}\bigr)
  +\lambda_{\text{prog}}\,
  \mathcal{L}_{\text{prog}}\bigl(p_t,\,\hat{p}_t\bigr)
\Bigr]
\label{eq:finetune_loss_final}
\end{split}
\end{equation}
Here, $\hat{a}_t = g_\psi(z_{\le t}, s_{\le t}, l)_{\text{[ACT]}}$ is the predicted action. $\mathcal{L}_{\text{act}}$ is the action loss combining Smooth L1 (joint regression), BCE (gripper command), and KL divergence (distribution regularization). $\hat{z}_{t+1:t+n} = g_\psi(z_{\le t}, s_{\le t}, l)_{\text{[OBS]}}$ are predicted future ViSA-Flow states, and $\mathcal{L}_{\text{obs}}$ is the forward dynamics loss (MSE, identical form to Eq.~\ref{eq:pretrain_loss_final} but on $D_{\tau}$) weighted by $\lambda_{\text{fwd}}$. $\hat{p}_t$ is the optional predicted progress, with $\mathcal{L}_{\text{prog}}$ being the progress loss (e.g., MSE) weighted by $\lambda_{\text{prog}}$. This stage adapts the general dynamics prior to the specific robot and learns the mapping from ViSA-Flow states (and proprioception) to robot actions, yielding the final policy parameters $\psi$.

\section{Evaluation}
\label{sec:evaluation}

\begin{figure*}[t!]
    \centering
    \includegraphics[width=1.0\linewidth]{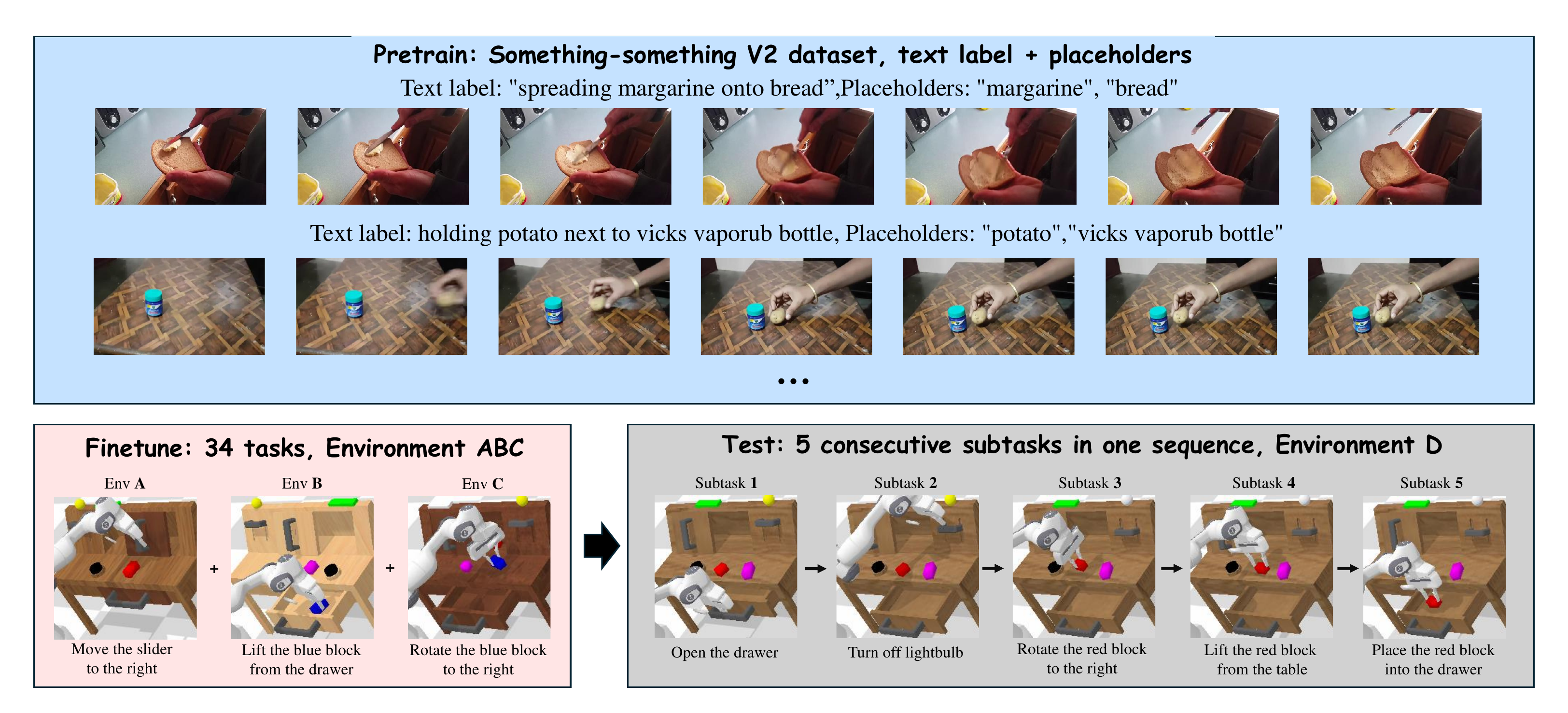}
      \vspace{-6mm}\caption{\textbf{Datasets used for pretraining, finetuning, and evaluation.} We pretrain on Something-Something-V2 with text labels and placeholders to extract semantic action flow. We finetune on 34 tasks across CALVIN environments A–C and evaluate zero-shot on environment D \cite{mees2022calvin}, where the robot completes 5 consecutive subtasks in one sequence.}
      \vspace{0.3cm}
      \label{fig: dataset}
      \vspace{-\baselineskip}
\end{figure*}

We conduct extensive experiments in both simulated and real-world environments to systematically evaluate ViSA-Flow's performance. Our evaluation is designed to answer the following key questions: 1) \label{q1} Can ViSA-Flow generalize across tasks with distractors, different backgrounds, and novel objects? 2) Can ViSA-Flow effectively learn and generalize across diverse tasks when expert demonstration data with language annotations are scarce? 3) Do semantic actions from human videos benefit robot skill learning?



\subsection{Simulation Experiments}


\noindent\textbf{Evaluation Setup.} We evaluate ViSA-Flow on the CALVIN benchmark\cite{mees2022calvin}, a standard testbed for long-horizon, language-conditioned manipulation requiring generalization. We use the ABC$\rightarrow$D split, training on environments A, B, C and evaluating zero-shot on the unseen environment D as shown in the lower row of Fig.~\ref{fig: dataset}.

\noindent \textbf{Pre-training Data.} The ViSA-Flow model undergoes pre-training (Stage 1, Sec.~\ref{sec:learning_framework_architecture}) using the large-scale Something-Something-V2 (SthV2) dataset\cite{goyal2017somethingsomethingvideodatabase} as the source domain. SthV2 contains approximately 220k short videos of human-object interactions with template text labels (examples visualized in the upper row of Fig.~\ref{fig: dataset}). The videos are processed to extract ViSA-Flow representations which are used for the pre-training as described in Secs.~\ref{sec:visaflow_representation} and \ref{sec:learning_framework_architecture}.



\noindent \textbf{Fine-tuning Data.} Following pre-training, ViSA-Flow is fine-tuned (Stage 2, Sec.~\ref{sec:learning_framework_architecture}) specifically for the CALVIN environment \cite{mees2022calvin}. To evaluate performance under data scarcity, we utilize only \textbf{10\%} (1,768 trajectories) of the available language-annotated robot demonstrations from CALVIN's ABC dataset as our target domain dataset. Each trajectory consists of the language instruction and the sequence of robot states, observations, and actions. 
Robot observations are processed into ViSA-Flow representations using the  pipeline described in Sec.~\ref{sec:visaflow_representation}.


\begin{table*}[t]
\centering
\caption{Comparative evaluation on CALVIN ABC$\rightarrow$
D benchmark. Performance metrics include success rates for completing 1-5 consecutive tasks and average sequence length (Avg. Len). Methods in the top section use 100\% of training data, while methods in the bottom section use only 10\%. The robot executed 1,000 test sequences with five tasks each. \textbf{Bold} indicates best performance.}\vspace{0mm}
\label{tab:simulation_experiment_summary}
\resizebox{\textwidth}{!}{ 
\begin{tabular}{l|c|c|ccccc|c}
\toprule

\multirow{2}{*}{\textbf{Method}} & \multirow{2}{*}{\makecell{\textbf{Fully-Annotated} \\ \textbf{Data (Demo No.)}}} & \multirow{2}{*}{\makecell{\textbf{Partially-Annotated} \\ \textbf{Data}}} & \multicolumn{5}{c|}{\textbf{Tasks Completed in A Row}} & \multirow{2}{*}{\textbf{Avg}. \textbf{Len}.} \\
\cmidrule(lr){4-8}
& & & \textbf{1} & \textbf{2} & \textbf{3} & \textbf{4} & \textbf{5} & \\
\midrule
\textcolor{gray}{Hulc \cite{mees2022matterslanguageconditionedrobotic}} & \textcolor{gray}{100\% (17870)} & \textcolor{gray}{\checkmark} & \textcolor{gray}{41.8\%} & \textcolor{gray}{16.5\%} & \textcolor{gray}{5.7\%} & \textcolor{gray}{1.9\%} & \textcolor{gray}{1.1\%} & \textcolor{gray}{0.67}\\
\textcolor{gray}{MDT \cite{reuss2024multimodaldiffusiontransformerlearning}} & \textcolor{gray}{100\% (17870)} & \textcolor{gray}{\checkmark} & \textcolor{gray}{61.7\%} & \textcolor{gray}{40.6\%} & \textcolor{gray}{23.8\%} & \textcolor{gray}{14.7\%} & \textcolor{gray}{8.7\%} & \textcolor{gray}{1.54}\\
\textcolor{gray}{Spil \cite{zhou2024languageconditionedimitationlearningbase}} & \textcolor{gray}{100\% (17870)} & \textcolor{gray}{\checkmark} & \textcolor{gray}{74.2\%} & \textcolor{gray}{46.3\%} & \textcolor{gray}{27.6\%} & \textcolor{gray}{14.7\%} & \textcolor{gray}{8.0\%} & \textcolor{gray}{1.71}\\
\textcolor{gray}{Roboflamingo \cite{li2024visionlanguagefoundationmodelseffective}} & \textcolor{gray}{100\% (17870)} & \textcolor{gray}{\xmark} & \textcolor{gray}{82.4\%} & \textcolor{gray}{61.9\%} & \textcolor{gray}{46.6\%} & \textcolor{gray}{33.1\%} & \textcolor{gray}{23.5\%} & \textcolor{gray}{2.47}\\
\textcolor{gray}{SuSIE \cite{black2023susie}} & \textcolor{gray}{100\% (17870)} & \textcolor{gray}{\checkmark} & \textcolor{gray}{87.0\%} & \textcolor{gray}{69.0\%} & \textcolor{gray}{49.0\%} & \textcolor{gray}{38.0\%} & \textcolor{gray}{26.0\%} & \textcolor{gray}{2.69}\\
\midrule
ATM \cite{wen2024anypointtrajectorymodelingpolicy}            & 10\% (1768) & \xmark & 31.7\% & 5.1\% & 1.3\% & 0.0\% & 0.0\% & 0.43\\ 
CLOVER \cite{bu2024closedloopvisuomotorcontrolgenerative}            & 10\% (1768) & \xmark & 44.3\% & 18.0\% & 5.0\% & 1.0\% & 0.0\% & 0.68\\ 
GR-1  \cite{wu2023unleashinglargescalevideogenerative}            & 10\% (1768) & \xmark & 67.2\% & 37.1\% & 19.8\% & 10.8\% & 6.9\% & 1.41 \\ 
SeeR  \cite{tian2024predictiveinversedynamicsmodels}            & 10\% (1768) & \xmark & 65.5\% & 38.8\% & 21.4\% & 11.7\% & 6.8\% & 1.44 \\ 
GR-MG  \cite{li2024grmg}          &10\% (1768) & \xmark & 81.8\% & 59.0\% & 39.0\% & 24.0\% & 16.2\% & 2.20 \\ 
\rowcolor{gray!25} ViSA-Flow (Ours)     & 10\% (1768) & \xmark & \textbf{89.0\%} & \textbf{73.8}\% & \textbf{56.8}\% & \textbf{44.8\%} & \textbf{31.4\%} &  \textbf{2.96}\\ 
\bottomrule
\end{tabular}
}
\vspace{-3mm}
\end{table*}

\noindent \textbf{Baselines.}We compare ViSA-Flow against two groups of SOTA methods: \textit{(i) Low-Data Baselines} trained on the same 10\% split for a fair data-efficiency comparison—ATM~\cite{wen2024anypointtrajectorymodelingpolicy}, a flow-based interface enabling any-point querying; CLOVER~\cite{bu2024closedloopvisuomotorcontrolgenerative}, a generative closed-loop visuomotor controller; GR-1~\cite{wu2023unleashinglargescalevideogenerative}, a multimodal transformer pre-trained on human videos; SeeR~\cite{tian2024predictiveinversedynamicsmodels}, a predictive inverse-dynamics approach; and GR-MG~\cite{li2024grmg}, a closely related transformer that augments GR-1 with explicit goal conditioning. We chose these baselines because they span complementary representations (flow-based, generative modeling, inverse dynamics, language/goal-conditioned transformers) and include models both with and without human-video pretraining, allowing us to isolate the effect of cross-domain priors. \textit{(ii) Full-Data Baselines:} Methods trained on 100\% of CALVIN annotated robot data (17,870 trajectories), including Hulc\cite{mees2022matterslanguageconditionedrobotic}, MDT\cite{reuss2024multimodaldiffusiontransformerlearning}, Spil\cite{zhou2024languageconditionedimitationlearningbase}, Roboflamingo\cite{li2024visionlanguagefoundationmodelseffective} and SuSIE\cite{black2023susie}. These represent the performance achievable with substantially more in-domain supervision.


\noindent\textbf{Metrics.} Following the standard CALVIN evaluation protocol\cite{mees2022calvin}, we measure the success rate to complete 5 consecutive subtasks, evaluated over 1,000 independent sequences. We also report the average successful sequence length (Avg. Len.). These metrics assess single-task proficiency and the ability to maintain performance over long horizons.

\begin{figure*}[t!]
    \centering
    \vspace{-0.0cm}
    \includegraphics[width=\linewidth]{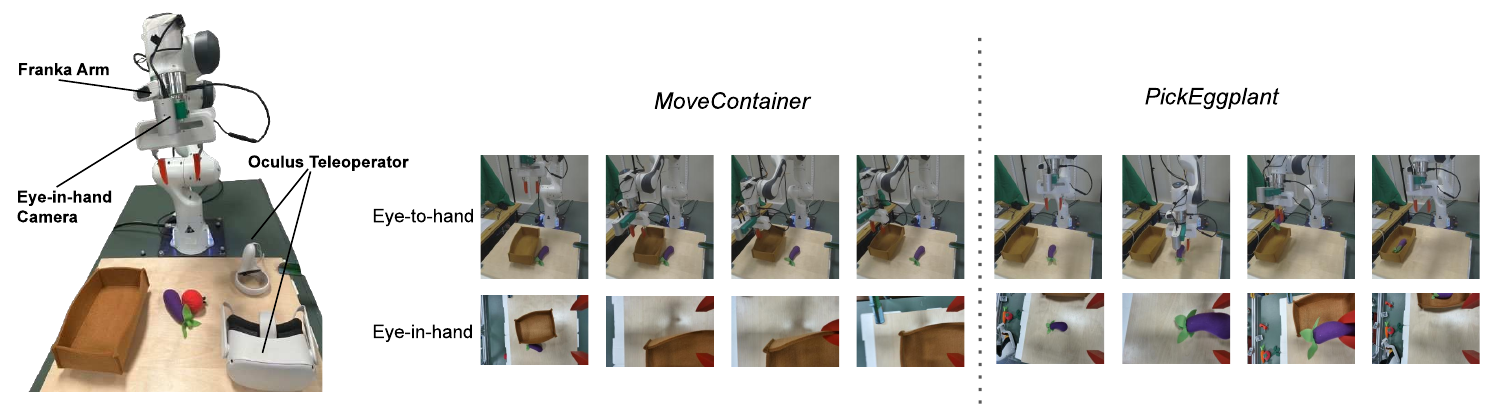}\vspace{-0mm}
      \caption{\textbf{The real-world experiment setup.}
      We evaluate ViSA-Flow on two single-stage manipulation tasks and a two-stage long-horizon manipulation task.
      }
      \vspace{-0.0cm}
      \label{fig: real_robot_setup}
      \vspace{-\baselineskip}
\end{figure*}

\noindent
\textbf{Results and Analysis.} Table~\ref{tab:simulation_experiment_summary} presents the performance metrics for all methods. The results demonstrate that ViSA-Flow outperforms all baseline methods, achieving highest success rates across all consecutive task completion metrics despite using only 10\% of the available annotated robot trajectories. Most impressively, ViSA-Flow maintains strong performance in sequential tasks, completing 5 consecutive tasks 31.4\% of the time, almost twice the rate of the next best method trained with 10\% data (GR-MG: 16.2\%) and exceeding all methods trained on 100\% data, including Susie (26.0\%). The average sequence length of 2.96 further demonstrates the effectiveness of ViSA-Flow in handling long-horizon manipulation tasks. Performance degradation from single to sequential tasks (89.0\% → 31.4\%) is notably less severe for ViSA (64.7\% reduction) compared to GR-MG (80.2\% reduction) and Susie (70.1\% reduction). This remarkable performance can probably be attributed to
utilization of semantic action representations extracted from human demonstration videos.
These results in simulation experiments validate our hypothesis that semantic action representations from human videos can significantly enhance robot skill learning, even when expert demonstrations are scarce and encounter different environments.


\begin{figure*}[htbp!]
    \centering
    \vspace{-0.cm}
    \includegraphics[width=\linewidth]{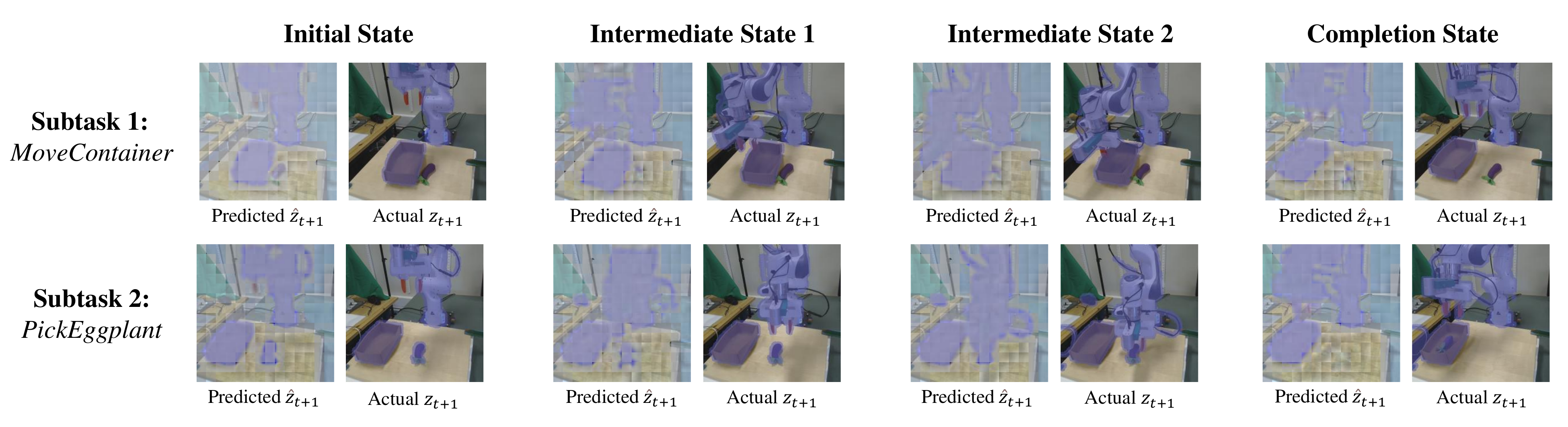}
      \vspace{-0mm}\caption{\textbf{Qualitative results on the real world long-horizon task.}  
    We visualize the \emph{decoded} ViSA-Flow prediction at ${\hat{z}}_{t+1}$ against the \emph{actual} ViSA-Flow ${z}_{t+1}$ extracted from the next observation for four execution phases.  
    Two rows correspond to the two subtasks that make up the long-horizon evaluation: \textbf{(Top)} \emph{Subtask 1 – MoveContainer;}  
    \textbf{(Bottom)} \emph{Subtask 2 – PickEggplant.} 
    Qualitatively, the model’s one-step predictions closely follow the true motion of the manipulator and task-relevant objects, even as the scene evolves across distinct interaction stages.}
      \vspace{-0.cm}
      \label{fig:qualitative}
\end{figure*}

\begin{figure}[t]          
  \centering
  \includegraphics[width=.48\textwidth]{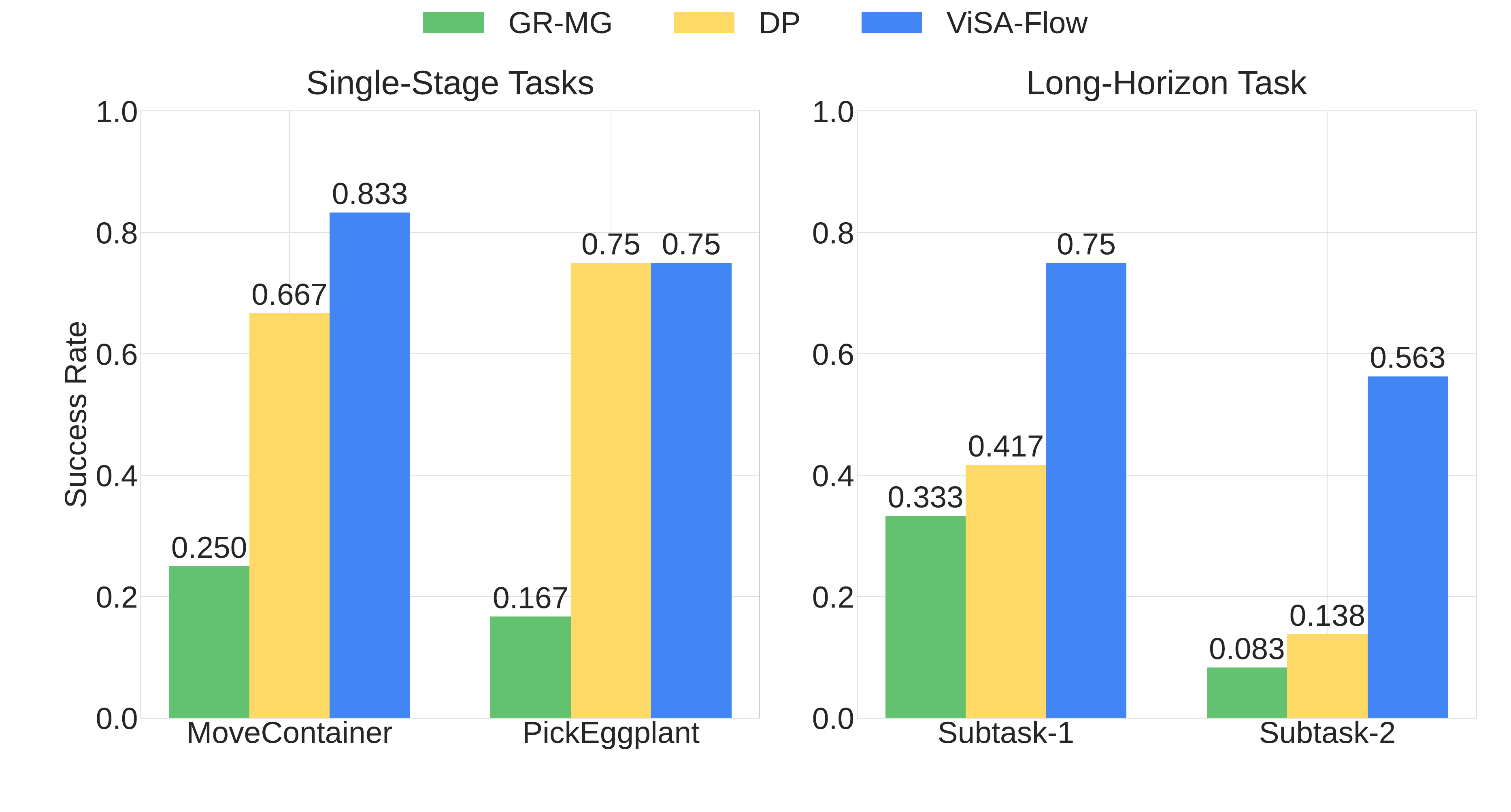}
  \vspace{-3mm}
  \caption{\textbf{Real‑world experimental results.} \textbf{Left:} two single‑stage tasks; \textbf{Right:} a two‑stage long‑horizon task.}
  \label{fig:real_world_results}
  \vspace{-0.cm}
\end{figure}

\noindent

\noindent
\textbf{Ablation Study of ViSA-Flow Components.} 
Table~\ref{tab:ablation} summarizes the results when each component within the ViSA-Flow framework is individually removed from the full method. Critically, removing the human-video pre-training stage (\textbf{w/o pre.}) leads to a near collapse in performance, indicating that the dynamics prior distilled from large-scale human videos is essential for multi-step task success. Removing semantic entity grounding (\textbf{w/o Seg.}, with $\alpha{=}0$) and tracking motion over whole images significantly reduces performance across all consecutive-task metrics: success on five-task sequences drops from 31.4\% to 9.6\%, and average successful length falls from 2.96 to 1.64, which indicates the importance of accurately segmenting and identifying semantic entities to anchor tracking and flow conditioning. Omitting temporal tracking (\textbf{w/o Trace.}) decreases average successful length from 2.96 to 2.78, highlighting that consistent point correspondences preserve temporal dynamics. Excluding manipulator grounding (\textbf{w/o Hand}) yields a modest drop (2.96 to 2.83) that validates segmentation and tracking are primary drivers while manipulator cues still aid spatial context. Overall, the full ViSA-Flow—integrating segmentation, tracking, and manipulator grounding—achieves the best results, and abtion results confirm that each component contributes to reliable long-horizon, cross-domain execution.

\begin{table}[t]
\centering
\caption{Ablation study evaluating the contribution of key components in ViSA-Flow.}\vspace{1mm}
\label{tab:ablation}
\resizebox{0.48\textwidth}{!}{
\begin{tabular}{l|ccccc|c}
\toprule
\multirow{2}{*}{\textbf{Method}} & \multicolumn{5}{c|}{\textbf{Tasks Completed in A Row}} & \multirow{2}{*}{\textbf{Avg. Len.}} \\
\cmidrule(lr){2-6}
& \textbf{1} & \textbf{2} & \textbf{3} & \textbf{4} & \textbf{5} \\
\midrule
ViSA-Flow w/o pre.           & 16.0\%  & 1.6\% &  0.0\% & 0.0\% &  0.0\%&  0.18\\
ViSA-Flow w/o Seg.           & 71.3\%  & 45.1\% &  24.5\% & 14.5\% &  9.6\%&  1.64\\
ViSA-Flow w/o Trace.        & 87.2\%  & 69.2\% &  52.0\% & 39.6\% &  30.0\%&  2.78\\
ViSA-Flow w/o Hand          &  89.0\%&  71.8\%&  54.2\%&  39.4\%&  28.4\%&  2.83\\
\rowcolor{gray!25} ViSA-Flow (Full)  & \textbf{89.0\%} & \textbf{73.8\%} & \textbf{56.8\%} & \textbf{44.8\%} & \textbf{31.4\%} & \textbf{2.96} \\
\bottomrule
\end{tabular}
} 
\vspace{-3mm}
\end{table}

\begin{table}[ht]
  \centering  \vspace{-2mm}
  \caption{Ablation study evaluating ViSA-Flow data scaling.} \vspace{0mm}
  \label{tab:data_ablation}
  \resizebox{0.48\textwidth}{!}{%
    \begin{tabular}{l|ccccc|c}
      \toprule
      \multirow{2}{*}{\textbf{Data}}
        & \multicolumn{5}{c|}{\textbf{Tasks Completed in a Row}}
        & \multirow{2}{*}{\textbf{Avg.\ Len.}} \\     
      \cmidrule(lr){2-6}
        & \textbf{1} & \textbf{2} & \textbf{3} & \textbf{4} & \textbf{5} & \\   
      \midrule
      5\% ABC$\!\rightarrow\!$D  & 84.6\% & 59.6\% & 41.0\% & 27.6\% & 18.4\% & 2.31 \\
      10\% ABC$\!\rightarrow\!$D & 89.0\% & 73.8\% & 56.8\% & 44.8\% & 31.4\% & 2.96 \\
      50\% ABC$\!\rightarrow\!$D & \textbf{93.8}\% & \textbf{85.4}\% & \textbf{76.2}\% & \textbf{68.8}\% & \textbf{58.8}\% & \textbf{3.83} \\
      \bottomrule
    \end{tabular}%
  }\vspace{-3mm}
\end{table}

\textbf{Data Scaling.} In addition, we evaluate ViSA-Flow under different amounts of robot demonstration data (5\%, 10\%, and 50\%). As shown in Table~\ref{tab:data_ablation}, performance consistently improves as the amount of data increases, highlighting ViSA-Flow’s scalability and data efficiency.

\subsection{Real World Experiments}
We evaluate the performance of ViSA-Flow in real-world experiments across diverse settings, focusing on its effectiveness and robustness in solving both single-stage and long-horizon tasks.

\noindent
\textbf{Experiment Setup.}
We evaluate our ViSA-Flow method in two real-world settings: two single-stage manipulation tasks and one long-horizon manipulation task. The demonstrations were collected by teleoperating a 7-DOF Franka Emika Panda arm using the Oculus-based application. We use two cameras (one eye-in-hand, one eye-to-hand) to provide RGB observations. The real-world experiment setup is shown in Fig.~\ref{fig: real_robot_setup}. For single-stage tasks, we collected 46 and 54 demonstrations for two tasks—\textit{MoveContainer} and \textit{PickEggplant} respectively. We train the ViSA-Flow policy for each single-stage task.
For long-horizon tasks, we consider the same two subtasks, \textit{MoveContainer} and \textit{PickEggplant}, requiring the robot to complete the first task before sequentially solving the second. This setup ensures consistency with the testing scenario used in our simulation experiments.
We pre-train our ViSA-Flow model on an NVIDIA RTX 4090 for 30 epochs, followed by fine-tuning for 30 epochs on single-stage tasks and 50 epochs for long-horizon tasks, respectively. 
We evaluate each policy across 12 different initial positions.

\noindent
\textbf{Baselines.} We compare our ViSA‑Flow method with GR‑MG \cite{li2024grmg} and the visuomotor Diffusion Policy (DP) \cite{chi2023diffusionpolicy}, both of which leverage RGB and proprioceptive inputs. To ensure fair comparison, all baseline models are trained on the same real‑world demonstration datasets for the two single‑stage tasks and the long‑horizon task.

\noindent
\textbf{Quantitative Results and Analysis.} The real-world experimental results are presented in Fig.~\ref{fig:real_world_results}. For the single-stage tasks \textit{MoveContainer} and \textit{PickEggplant}, ViSA-Flow significantly outperforms the GR-MG model across 12 trials. Meanwhile, DP achieves a comparable success rate of 75.0\% on the \textit{PickEggplant} task. In contrast, for the long-horizon task—which sequentially combines \textit{MoveContainer} and \textit{PickEggplant}—our method demonstrates superior performance, achieving 9/12 successful trials for each subtask and yielding an overall success rate of 56.3\% for the full sequence. By comparison, GR-MG and DP attain success rates of only 8.3\% and 13.8\%, respectively. Notably, DP experiences a significant performance drop when transitioning from single-stage to long-horizon tasks, whereas ViSA-Flow maintains robust and consistent performance.

\noindent
\textbf{Qualitative Results and Analysis.} Fig.~\ref{fig:qualitative} qualitatively demonstrates that the decoded ViSA-Flow one-step prediction $\hat{z}_{t+1}$ remains tightly aligned with the ground-truth flow throughout the entire long-horizon execution: the model persistently focuses on the robot gripper and the task-relevant objects while suppressing background clutter, its spatial support evolves smoothly and coherently as the scene transitions from the initial approach, through two intermediate contact phases, to the completion state, and the same level of accuracy is observed across the two sequential subtasks.
This close match between prediction and observation confirms that the cross-domain dynamics prior learned during pretraining effectively captures task-critical interaction structure and generalizes to novel real-world embodiments.

\section{Limitations and Future Work}
\label{sec:limitation}
While ViSA-Flow demonstrates strong performance in observational robot learning, it currently lacks explicit modeling of 3D geometry and contact dynamics, which may limit its generalization to tasks requiring fine-grained physical interactions. The framework also relies on pretrained VLM components, potentially restricting adaptability to novel domains or unseen objects. Moreover, ViSA-Flow currently transfers only object–manipulator interactions from human manipulation videos to robots, ignoring more nuanced dexterous movements of human fingers. This simplification leads to some loss of manipulation knowledge, which may limit performance in tasks requiring fine or dexterous control.

Future work aims to address these limitations by enriching ViSA-Flow representations with contact physics and 3D reasoning, reducing reliance on pretrained models through joint or end-to-end training with VLMs, and integrating its learned priors with reinforcement learning to enhance policy learning. Additionally, we plan to investigate methods to capture and transfer finer-grained human manipulation skills, preserving dexterous finger-level knowledge for robotic use. Scaling pretraining to large-scale video corpora and further analyzing ViSA-Flow’s invariance properties and sample efficiency also represent promising directions for advancing robust and generalizable robot learning from observation.
\bibliographystyle{IEEEtran}
\bibliography{references}

\end{document}